\documentclass[conference]{IEEEtran}
\IEEEoverridecommandlockouts
\usepackage{cite}
\usepackage{amsmath,amssymb,amsfonts}
\usepackage{algorithmic}
\usepackage{graphicx}
\usepackage{textcomp}
\usepackage{xcolor}
\def\BibTeX{{\rm B\kern-.05em{\sc i\kern-.025em b}\kern-.08em
    T\kern-.1667em\lower.7ex\hbox{E}\kern-.125emX}}

\usepackage{color}
\usepackage{subcaption}
\usepackage{verbatim}
\usepackage{pdflscape}
\usepackage[T1]{fontenc}
\usepackage{listings}
\usepackage{listingsutf8}

\lstdefinelanguage{PRISM}{
    keywords={dtmc, const, module, endmodule, rewards, endrewards, label, true, false, bool, double, init, []},
    keywordstyle=\color{blue}\bfseries,
    morecomment=[l]{//},    
    commentstyle=\color{green!50!black}\itshape,
    morestring=[b]",        
    stringstyle=\color{red}\ttfamily,
    sensitive=true,
    morekeywords={const, double, true, false}, 
}

\begin{document}

\title{Symbolic Runtime Verification and Adaptive Decision-Making for Robot-Assisted Dressing
\thanks{We are grateful to the EPSRC for support (UKRI Trustworthy Autonomous Systems (TAS) Node in Resilience, grant  EP/V026747/1; TAS Node in Verifiability, EP/V026801/2). We are grateful to The North Yorkshire County Council. }
}

\author{\IEEEauthorblockN{Yasmin Rafiq}
\IEEEauthorblockA{\textit{Department of Computer Science} \\
\textit{University of Manchester}, UK \\
yasmeen.rafiq@manchester.ac.uk}
\and
\IEEEauthorblockN{Gricel  V\'{a}zquez}
\IEEEauthorblockA{\textit{Department of Computer Science} \\
\textit{University of York}, UK \\
gricel.vazquez@york.ac.uk}
\and
\IEEEauthorblockN{Radu Calinescu}
\IEEEauthorblockA{\textit{Department of Computer Science} \\
\textit{University of York}, UK \\
radu.calinescu@york.ac.uk}
\and
\IEEEauthorblockN{Sanja Dogramadzi}
\IEEEauthorblockA{\textit{School of Electrical and Electronic Engineering} \\
\textit{The University of Sheffield}, UK \\
s.dogramadzi@sheffield.ac.uk}
\and
\IEEEauthorblockN{Robert M Hierons}
\IEEEauthorblockA{\textit{School of Computer Science} \\
\textit{The University of Sheffield}, UK \\
r.hierons@sheffield.ac.uk}
}

\maketitle


\begin{abstract}
We present a 
control framework for robot-assisted dressing that augments low-level hazard response with runtime monitoring and formal verification. A parametric discrete-time Markov chain (pDTMC) models the dressing process, while Bayesian inference dynamically updates this pDTMC's transition probabilities based on sensory and user feedback. Safety constraints from hazard analysis are expressed in probabilistic computation tree logic, and symbolically verified using a probabilistic model checker. We evaluate reachability, cost, and reward trade-offs for garment-snag mitigation and escalation, enabling real-time adaptation. Our approach provides a formal yet lightweight foundation for safety-aware, explainable robotic assistance.
\end{abstract}


\section{Introduction}

As assistive robots become more integrated into daily life, ensuring safety and reliability during physical human-robot interaction (pHRI) remains a critical challenge~\cite{christoforou2020upcoming,cooper2020ari,erickson2020assistive,gu2021major}. In tasks such as Robot-Assisted Dressing (RAD)~\cite{jevtic2018personalized,chance2017quantitative}, the robot must perform close-contact assistance in coordination with a human, where garment dynamics, force disturbances, and user motion introduce runtime uncertainty.
While low-level control strategies (e.g., force thresholds, compliant motions, and speed modulation) can mitigate immediate issues,
they operate reactively and lack long-term task-level reasoning. This motivates the need for high-level control strategies that reason over future states and adapt using runtime feedback.


We propose a high-level control framework that integrates parametric discrete-time Markov chains (pDTMCs)~\cite{hahn2010param}, Bayesian inference~\cite{calinescu2014adaptive}, and symbolic runtime verification~\cite{kwiatkowska2004probabilistic}. Safety and performance requirements, derived from a hazard analysis~\cite{delgado2021safety}, are formally specified using probabilistic computation tree logic (PCTL)~\cite{ciesinski2004probabilistic}. Symbolic expressions for these requirements are precomputed using the probabilistic model checkers PRISM~\cite{KNP11} and PARAM~\cite{hahn2010param}, and are evaluated at runtime using parameters updated via Bayesian inference. 

The main contributions of our paper are:
\begin{itemize}
    \item A high-level formal verification framework using a pDTMC model guided by hazard analysis.
    \item Integration of runtime Bayesian inference to update model parameters using sensory data and user feedback.
    \item Symbolic runtime verification using closed-form expressions generated with PRISM and PARAM.
    \item Symbolic evaluation of safety, cost, and reward properties to support dynamic adaptation and decision-making.
\end{itemize}

\noindent
The rest of the paper is structured as follows. Section~\ref{sec:relatedwork} reviews related work. Section~\ref{sec:systemoverview} describes the system architecture. Section~\ref{sec:verification} defines the pDTMC model and its formal requirements, and Section~\ref{sec:evaluation} presents its symbolic analysis. Section~\ref{sec:discussion} discusses practical insights and limitations. Section~\ref{sec:conclusion} summarises our results and outlines future work directions.


\section{Related Work}
\label{sec:relatedwork}

This section reviews research at the intersection of probabilistic verification, runtime adaptation, and uncertainty quantification in robotics, particularly under pHRI settings.

\paragraph{Probabilistic Verification and Model Checking}
Probabilistic model checking has been widely applied in robotics to provide formal guarantees over safety, reachability, and performance requirements under uncertainty. Tools such as PRISM~\cite{KNP11} support the analysis of Discrete-Time Markov Chains (DTMCs) and verification of requirements specified in PCTL~\cite{ciesinski2004probabilistic}.

Zhao et al.~\cite{zhao2019probabilistic} applied probabilistic verification to autonomous systems in extreme environments, but their approach focused on offline analysis without runtime adaptation. Similarly, Gleirscher et al.~\cite{gleirscher2020safety, gleirscher2022verified} verified safety controllers for collaborative robots but did not incorporate runtime feedback or symbolic reasoning in human-interactive scenarios.

\paragraph{Bayesian Inference for Uncertainty Quantification}
Bayesian learning has been used to adapt robot behaviour based on noisy perception and dynamic environments. Zhao et al.~\cite{zhao2024bayesian} proposed a framework that combines Bayesian inference with formal verification to improve robustness in autonomous systems. Similarly, Calinescu et al. present a tool for the verification and Bayesian-learned parameter inference of probabilistic world models~\cite{calinescu2025verification}. However, they did not consider symbolic runtime evaluation or human-in-the-loop feedback. 

\paragraph{Runtime Monitoring and Adaptive Control}
Runtime monitoring techniques are often used for anomaly detection or logging, whereas runtime verification aims to formally check execution traces against temporal specifications. Kirca et al.~\cite{kirca2023runtime} proposed a runtime monitoring framework for anomaly detection using logs, but without safety property verification. Li et al.~\cite{li2021provably} introduced a probabilistic motion planning approach that models human uncertainty, but it lacks continuous verification of safety requirements during task execution.

\textit{Our Contribution in Context:} 
In contrast to the above, our work integrates probabilistic model checking, runtime Bayesian inference, and hazard-driven safety property verification in a human-in-the-loop context. We introduce a pDTMC model whose transition probabilities are updated at runtime using Bayesian learning. Precomputed symbolic expressions for PCTL properties are evaluated on-the-fly through parameter substitution, enabling lightweight runtime verification without re-invoking the model checker. 
Our framework bridges the gap between high-assurance probabilistic verification and adaptive human-aware control. It enables formal safety reasoning in real time while responding to user input, physical interaction events and task-level uncertainty during robot-assisted dressing.


\section{System Overview and Modelling Framework}\label{sec:systemoverview}

\subsection{Robot-Assisted Dressing Context}
\label{subsec:rad_context}

Robot-Assisted Dressing (RAD) involves a robotic manipulator physically assisting a user in donning garments (e.g., jacket). The task is inherently collaborative, requiring safe human-robot interaction, real-time trajectory execution, and continuous hazard monitoring to ensure comfort and prevent failure. 
Dynamic factors such as user movement, garment deformation, and variable force profiles can lead to hazards including garment snagging or user discomfort. A robust RAD system must detect such events and respond appropriately.

In our setup, a validated low-level control system is responsible for executing motion trajectories, monitoring contact forces, and responding to user-issued verbal feedback (e.g., reports of pain). These behaviours will be detailed in a companion paper on reactive control strategies.

This paper focuses on the high-level control framework that operates above the low-level controller. It models the dressing task as a probabilistic discrete-time Markov chain (pDTMC), enabling symbolic runtime verification of safety and performance requirements. The high-level system continuously updates beliefs about task state using Bayesian inference, substitutes these into precomputed symbolic expressions, and adaptively refines robot decision-making. 

Together, this layered architecture supports both proactive and reactive responses to runtime uncertainty, advancing safe and adaptive human-robot collaboration in dressing assistance.

\subsection{Low-Level Control Strategy overview}

The low-level control strategies implemented in our RAD system serve as the foundational layer for real-time hazard mitigation. These strategies were developed and validated through physical dressing trials involving simulated garment snags and user-reported discomfort. Full details will be provided in a companion study.
The two primary strategies are:

\begin{itemize} 
\item \textbf{Garment Snagging Control:} When excessive interaction forces are detected via integrated force sensors, the system triggers one of three responses: (a) prompt the user for assistance through a Rasa-based chatbot interface, (b) initiate autonomous recovery by adjusting the robot's trajectory, or (c) abort the task safely if mitigation fails.

\item \textbf{User Discomfort Mitigation:} Enables users to express pain or discomfort through natural language commands. The robot responds by progressively reducing speed and, if needed, aborting the task.
\end{itemize}

These low-level mechanisms are essential for reactive safety during dressing and were experimentally validated across multiple dressing trials, showing effective mitigation of hazards in both user-assisted and autonomous recovery modes.

The high-level control strategies introduced in this work are designed to complement these reactive behaviours by offering predictive, symbolic reasoning capabilities through pDTMC-based runtime verification. Together, the combined architecture ensures both proactive and reactive safety handling in robot-assisted dressing. 

\subsection{High-Level Control Architecture}
\label{subsec:highlevel_architecture}

\begin{figure}
    \centering
    \includegraphics[width=0.4\textwidth]{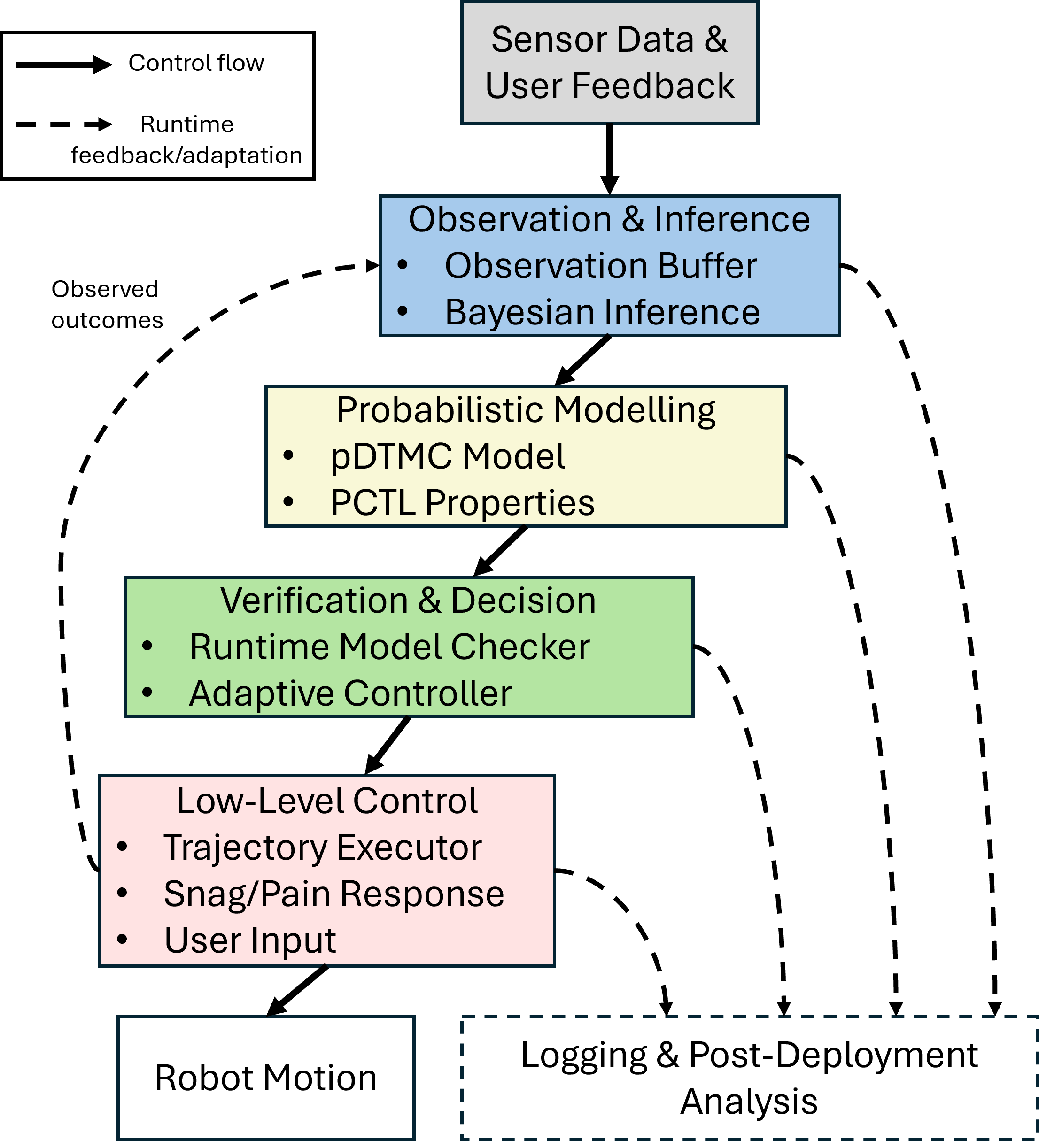}
    \caption{High-level control architecture showing runtime adaptation via symbolic verification.}
    \label{fig:system_architecture}

\end{figure}

Figure~\ref{fig:system_architecture} illustrates the runtime-adaptive control architecture for the RAD
system. The architecture integrates low-level sensor processing with high-level symbolic reasoning to support safe and responsive operation under uncertainty.

At the top of the pipeline, the system receives continuous inputs from sensors (e.g., force, trajectory deviation) and verbal user feedback. These inputs are processed by an \textit{Observation and Inference} module that maintains a buffer of recent events and uses Bayesian inference with observation ageing~\cite{calinescu2011using,calinescu2014adaptive} to update runtime estimates of key transition probabilities (e.g., snag detection, recovery success).

These updated probabilities are injected into a symbolic \textit{Probabilistic Model} comprising a parametric Discrete-Time Markov Chain (pDTMC) and a set of PCTL-specified safety and reliability requirements. The model is evaluated by the \textit{Verification and Decision} layer using closed-form expressions obtained via \texttt{PRISM+PARAM}~\cite{hahn2010param}, enabling rapid runtime checks without re-invoking the model checker.

The output of this layer informs the \textit{Adaptive Controller}, which adjusts system behaviour based on the verification results. If a safety threshold is violated (e.g., escalation risk exceeds 0.5), the system proactively transitions to compliant or abortive behaviour. Otherwise, it proceeds with recovery or continues execution.

Finally, the \textit{Low-Level Control} layer executes trajectory plans, responds to snag and pain feedback, and integrates real-time user input. These behaviours are conditionally triggered based on the decisions verified in the upper layers. 
Together, this architecture ensures that the RAD system adapts at runtime to user-specific conditions while maintaining conformance with formal safety constraints. The pDTMC abstraction enables runtime introspection and supports explainability in decision-making.

\subsection{Hazard Analysis and Safety Constraints}
\label{subsec:hazard_analysis}

The pDTMC model and associated evaluation requirements were informed by a structured hazard analysis conducted during system design~\cite{delgado2021safety}. This analysis identified the following safety-critical scenarios related to task failure, user discomfort, and mitigation breakdowns, and derived constraints that the system must satisfy to maintain safe and reliable operation:


\begin{itemize}
    \item \textbf{Limit the risk of task abortion.} The system should minimise the likelihood of entering an abort state due to unresolved snags or escalation failures.
    
    \item \textbf{Ensure reliable task completion.} The dressing task should succeed in a high proportion of executions, even under varying user behaviour and environmental noise.
    
    \item \textbf{Bound the expected cost of silent failures.} Escalations that go undetected should not incur high expected penalties, motivating prompt detection and mitigation.

    \item \textbf{Avoid prolonged or delayed recovery.} Mitigation pathways should resolve issues efficiently, avoiding excessive retries or delays in user or autonomous responses.
\end{itemize}

\noindent
These constraints are later formalised in PCTL, and evaluated via symbolic expressions over transition parameters. This ensures that the RAD system maintains compliance with safety and comfort requirements—even in uncertain and dynamic execution contexts.


\section{
Snagging Model and Formal Verification}
\label{sec:verification}

\begin{figure}
    \centering
    \includegraphics[width=0.5\textwidth]{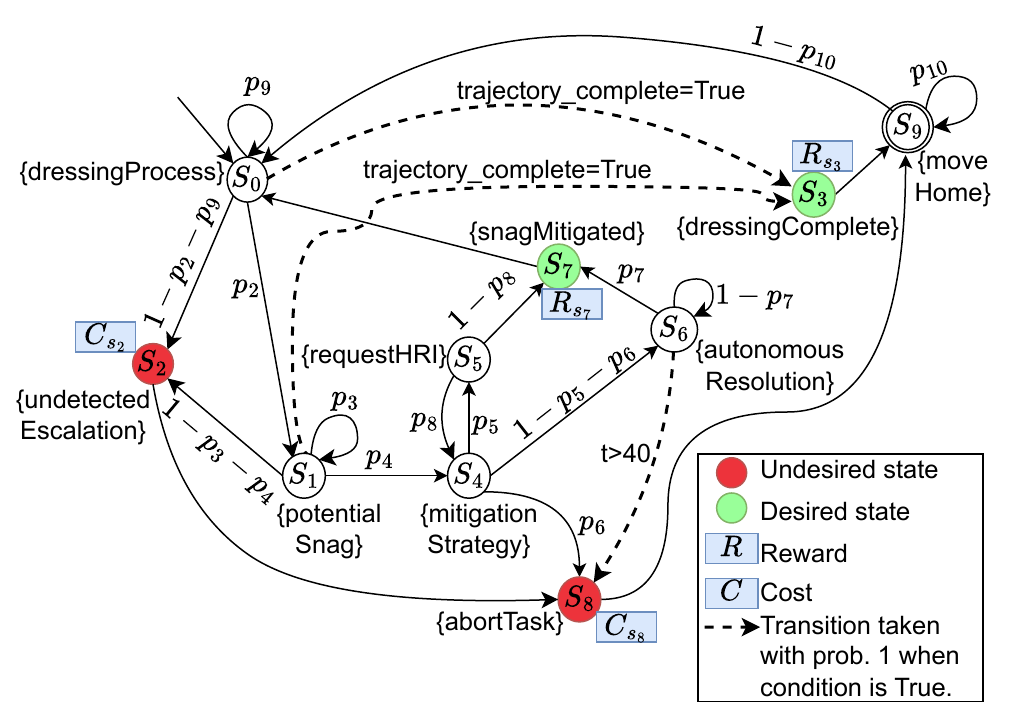}
    \caption{Abstraction of the pDTMC RAD model. Transition probabilities (\(p_{[.]}\)) govern state transitions. Timeout conditions (e.g., \(t = \texttt{MAX\_TIME}\) in \(s_6\)) ensure escalation resolution. Rewards (\(R_{[.]}\)) and costs (\(C_{[.]}\)) are linked to outcome states. 
    }
    \label{fig:snagDTMC}
\end{figure}

To verify safety-critical behaviour in robot-assisted dressing, we formalise the high-level control strategy as a pDTMC. This model captures task progression, garment-snag escalation, and recovery pathways as probabilistic transitions, abstracting from the robot’s low-level motion control.

\medskip\noindent
\textbf{Model description.} 
The pDTMC models 10 abstract task states (Fig.~\ref{fig:snagDTMC}). The initial dressing task begins in state \(s_0\) (\texttt{dressingProcess}), progressing towards \(s_3\) (\texttt{dressingComplete}). Potential snags are detected in \(s_1\) (\texttt{potentialSnag}) or escalated undetected to \(s_2\) (\texttt{undetectedEscalation}). If mitigation is required, the system enters \(s_4\) (\texttt{mitigationStrategy}), choosing between human assistance \(s_5\) (\texttt{requestHRI}) or autonomous resolution \(s_6\) (\texttt{autonomousResolution}). Successful mitigation reaches \(s_7\) (\texttt{snagMitigated}); failure results in task abortion \(s_8\) (\texttt{abortTask}). All outcomes eventually transition to \(s_9\) (\texttt{moveHome}) before resetting.

\medskip\noindent
\textbf{Formal definition.} The pDTMC is defined as a tuple $\mathcal{M} = (S, s_0, P, \mathcal{L}, \rho)$, where:
\begin{itemize}
    \item $S$ is the finite set of states. Each state is represented by:
    \[
    s' = (s,~t,~\texttt{time\_step},~\texttt{trajectory\_complete})
    \]
    where:
    \begin{itemize}
        \item \(s \in [0,9]\) is the current task stage.
        \item \(t \in [0, \texttt{MAX\_TIME}]\) tracks time in \(s_6\) (autonomous retries).
        \item \texttt{time\_step} tracks progress in \(s_0, s_1\) until trajectory completion.
        \item \texttt{trajectory\_complete} is a Boolean flag signalling task end.
    \end{itemize}

    \item \(s_0\) is the initial state: dressing process starts with all timers at zero.

    \item \(P: S \times S \rightarrow [0,1]\) is the parametric transition matrix governed by symbolic parameters \(p_1\) to \(p_{10}\), representing task observations and control uncertainties. All transition logic is defined by the PRISM model in Appendix~A.

    \item \(\mathcal{L}: S \rightarrow \mathcal{AP}\) labels an abstract state with atomic propositions (e.g., \texttt{dressingComplete}), enabling property specification such as \(P_{\leq 0.1}[F\, \texttt{abortTask}]\).

    \item \(\rho: S \rightarrow \mathbb{R}_{\geq 0}\) assigns rewards and costs to terminal states, capturing success and failure outcomes. These are:
    \begin{itemize}
        \item \(R_{s_3} = BASE\_REWARD\_S3 = 20\) - base reward constant for successful dressing (used in symbolic analysis).
        \item \(R_{s_7} = R\_S7 = 10\) for snag mitigation.
        \item \(C_{s_2} = C\_S2 = 10\) for undetected escalation.
        \item \(C_{s_8} = C\_S8 = 5\) for task abortion.
    \end{itemize}
\end{itemize}

\noindent In our symbolic analysis, $R_{s_3}$ is set to 20 as previously described; however, in the simulation, we further examine the effect of time by modelling this reward as a decayed value, as detailed below.

\smallskip\noindent
\textbf{Reward decay.}
Although cumulative reward properties are not symbolically supported by \texttt{PRISM+PARAM}, we model a decaying reward in the PRISM simulation model to discourage long delays in dressing completion:
\[
R_{s_3} = e^{-\texttt{DECAY\_RATE} \cdot \texttt{time\_step}} \cdot BASE\_REWARD\_S3
\]
with \texttt{DECAY\_RATE} set to 0.5. This decay reward is applied during simulation-based evaluation only and is excluded from the symbolic analysis in Section~\ref{sec:evaluation}, which uses the constant \texttt{BASE\_REWARD\_S3}.

\smallskip\noindent
\textbf{Model features.}
The pDTMC supports runtime verification and adaptive reasoning through:
\begin{itemize}
    \item \textbf{Symbolic transitions:} Parametric probabilities capture runtime uncertainty in detection, recovery, and escalation.
    \item \textbf{Progress tracking:} \texttt{trajectory\_complete}  and \texttt{time\_step} govern dressing task duration and allow symbolic evaluation of task timing.
    \item \textbf{Reward/cost abstraction:} Terminal states encode outcomes for reward-guided or risk-sensitive control.
    \item \textbf{Time-bounded retries:} Timeout in \(s_6\) prevents infinite loops in autonomous recovery.
\end{itemize}

This model provides the formal basis for the symbolic evaluation in Section~\ref{sec:evaluation}, and is structured to support runtime adaptation via parameter updates (see Discussion, Section~\ref{sec:discussion}).


\subsection{Specification of Safety and Reliability Requirements}
\label{subsec:pctl_spec}

The safety and performance requirements identified through hazard analysis (Section~\ref{subsec:hazard_analysis}) are formalised using Probabilistic Computation Tree Logic (PCTL) \cite{ciesinski2004probabilistic}. These requirements are used to evaluate the symbolic pDTMC model and guide runtime decision-making.

PCTL enables the specification of probabilistic reachability, time-bounded, and reward-bounded behaviours in discrete-time systems. For example, $P_{\leq 0.1}[F\, s=8]$ specifies that the probability of eventually aborting the task should be no greater than 10\%. Although \texttt{PRISM+PARAM} currently supports only reachability requirements in symbolic form, we approximate cost-based constraints using reachability proxies (see Section~\ref{sec:cost_analysis}).

Finally, the following requirements formalise the core constraints derived from the hazard analysis~\cite{delgado2021safety}: 

\begin{itemize}
    \item \textbf{(H1) Limit the risk of task abortion}
    \begin{equation}
    P_{\leq 0.1}[F\, s=8]
    \end{equation}
    ``The probability of reaching the task abort state $s=8$ must remain below 10\%.''

    \item \textbf{(H2) Ensure reliable task completion}
    \begin{equation}
    P_{\geq 0.9}[F\, s=3]
    \end{equation}
    ``The dressing task should complete successfully in at least 90\% of executions.''

    \item \textbf{(H3) Bound undetected escalation cost}
    \begin{equation}
     ExpectedCost_{s2} = C_{S2} \cdot P_{=?}[F\, s=2] \leq MAX\_C2
    \label{eq:ExpectedCost_Fs=2}
    \end{equation}
    ``The expected cost of undetected escalation (reaching $s=2$) must remain within limit set by the maximum cost allowed $MAX\_C2$.'' This is evaluated symbolically via a proxy expression.

    \item \textbf{(H4) Timely mitigation success reward}
    \begin{equation}
     ExpectedReward_{s7} = R_{S7} \cdot Pmax_{=?}[F\, s=7]
    \label{eq:ExpectedReward_Fs=7}
    \end{equation}
    ``The system should maximise the probability of successfully mitigating snags.'' The expected reward is derived symbolically based on $p_7$ and $p_8$.

    \item \textbf{(H5) Encourage time-bounded completion}
    \begin{equation}
    P_{\geq 0.95}[F_{\leq \texttt{MAX\_TIME\_TRAJ}}\, s=3]
    \end{equation}
    ``The dressing task should complete within the designated trajectory time in at least 95\% of cases.'' This is used in simulation.
\end{itemize}

These formal specifications provide a rigorous basis for symbolic evaluation and runtime adaptation. Requirements (H1)–(H4) are fully captured by symbolic expressions derived via \texttt{PRISM+PARAM}, enabling lightweight verification through parameter substitution. Time-bounded and cumulative reward constraints are approximated or evaluated using proxy expressions where symbolic support is unavailable.

\subsection{Symbolic Model Checking using PRISM+PARAM}
\label{subsec:prism-param}

We perform symbolic verification of the pDTMC model using the \texttt{PRISM+PARAM} toolchain \cite{hahn2010param}. This enables closed-form algebraic expressions to be extracted for PCTL reachability requirements such as $P_{=?}[F\, s=2]$ and $P_{=?}[F\, s=7]$. These symbolic expressions are parametrised over key transition probabilities (e.g., $p_2$, $p_3$, $p_7$, $p_8$), enabling analysis of system behaviour under uncertainty and variable conditions. 
Each expression captures the probability of reaching a safety-critical or goal state as a function of the system’s current configuration. In our evaluation (Section~\ref{sec:evaluation}), these expressions are visualised over bounded parameter ranges to assess risk and performance.

While the symbolic engine supports reachability queries, cumulative reward properties of the form $P_{=?}[C \leq T]$ are not currently supported. To address this, we reformulate cost and reward analysis using proxy expressions—multiplying the reachability probability by a fixed reward or cost scalar (e.g., see Equation~\ref{eq:ExpectedCost_Fs=2}).


\subsection{Runtime Verification using Bayesian Learning}
\label{subsec:runtime-verification}

To enable adaptive decision-making, the RAD system performs runtime verification by updating the transition probabilities of the pDTMC using Bayesian learning. This approach applies \emph{observation ageing} \cite{calinescu2011using,calinescu2014adaptive}, giving more weight to recent observations while discounting older data, allowing the system to respond to changes in user behaviour or task context.

The updated estimate of transition probability $p_{ij}^{(k)}$ after $k$ observations is computed as:

\begin{equation}
p_{ij}^{(k)} = \frac{c_0}{c_0 + k} p_{ij}^{(0)} + \frac{k}{c_0 + k} \cdot \frac{\sum_{l=1}^{k} w_l \cdot x_{ij}^{(l)}}{\sum_{l=1}^{k} w_l}
\end{equation}
\noindent
Here, $p_{ij}^{(0)}$ is the prior estimate, $c_0$ controls the influence of the prior, and $w_l = \alpha^{-(t_k - t_l)}$ applies exponential ageing to past observations. The decay factor $\alpha \geq 1$ controls how quickly older data are discounted.

The symbolic expressions derived offline for reachability and reward properties (e.g., $P_{=?}[F\, s=7]$) are evaluated at runtime by substituting dynamically updated probabilities (e.g., $p_7$, $p_8$). This enables lightweight, real-time runtime verification to guide decision-making during dressing without invoking a model checker.

\subsection{How the Adaptation Mechanism Operates}
\label{subsec:adaptation_mechanism}

During execution, the RAD system monitors real-time sensor data (e.g., garment force feedback, joint velocities) and user responses to update transition probabilities within the symbolic pDTMC model. These updated probabilities are substituted into precomputed symbolic expressions for key safety and performance requirements, allowing the system to assess evolving risk and make informed decisions at runtime.

If a safety threshold is violated—e.g., the probability of task abortion exceeds a bound—the system responds on two levels:

\begin{enumerate}
    \item \textbf{Low-Level Control Actions}: The controller immediately enters a compliant mode, reducing speed and applied force or halting movement to mitigate user discomfort or mechanical risks.
    
    \item \textbf{High-Level Adaptation}: The symbolic pDTMC model is used to track evolving task context. Bayesian learning updates transition probabilities (e.g., likelihood of snag, success of user or autonomous recovery), enabling dynamic substitution into symbolic expressions (e.g., $P_{=?}[F\, s=2]$). This supports real-time risk evaluation without rechecking the model.
\end{enumerate}

For example, if repeated snagging is observed under specific motion trajectories, the estimated probability of transitioning from $s_0$ (\texttt{dressingProcess}) to $s_1$ (\texttt{potentialSnag}) increases. Similarly, success rates of user-assisted or autonomous mitigation are reflected in the transition probabilities to $s_5$ and $s_6$. Rather than storing full execution histories, the Bayesian update loop maintains a compact belief over transition likelihoods.

This integration of symbolic runtime evaluation and adaptive control ensures that the system responds promptly to immediate safety threats while gradually improving high-level decision-making in dynamic, user-specific contexts.


\section{Evaluation}\label{sec:evaluation}

To assess the
robot-assisted dressing system under uncertainty, we adopt a symbolic evaluation approach using the \texttt{PRISM+PARAM} toolchain. Our evaluation focuses on three core aspects: (i) the reachability of critical failure or successful mitigation states, (ii) the expected cost of failure, and (iii) the expected reward of mitigation success.

Rather than relying on simulation-based numerical experiments, we symbolically evaluated the parametric DTMC model across key parameters that influence the system's decision-making behaviour—such as the reliability of snag detection, human and autonomous recovery, and escalation outcomes. These symbolic expressions allow for offline design analysis and enable runtime reasoning via parameter substitution. Each symbolic property is visualised using annotated heatmaps and contour plots for interpretability. 

\subsection{Symbolic Reachability Analysis of Snag Mitigation}\label{sub:eval-symbReachability}

We analysed the probability of successfully mitigating a snag ($s=7$) in the pDTMC model (Figure~\ref{fig:snagDTMC}) using \texttt{PRISM+PARAM}. The property of interest is the symbolic reachability expression $P=?[F\,s=7]$, which quantifies the probability of eventually reaching the successful mitigation state under varying system conditions. 

A closed-form symbolic expression was derived, capturing the joint influence of snag detection, escalation, and recovery through both autonomous and human intervention. To aid interpretability, we introduce the following terms:

\begin{itemize}
    \item $\alpha = p_2.p_4$ — snag detection and escalation,
    \item $\beta = p_5.p_8$ — failed recovery attempts via human mitigation,
    \item $\delta_k = p_7^k$ — recursive retries of autonomous recovery (limited to 40 steps).
\end{itemize}

A simplified excerpt of the symbolic expression extracted using \texttt{PRISM+PARAM} is shown below. This highlights the compound effects of detection ($\alpha = p_2 \cdot p_4$), human failure ($\beta = p_5 \cdot p_8$), and recursive autonomous retries ($\delta_k = p_7^k$):
\begin{equation}
P_{s=7} = \frac{
\begin{aligned}
& 10000\, \beta^2 \cdot \alpha^2 - 20000\, \beta \cdot p_5 \cdot \alpha^2 + \cdots + \\
& 100\, \beta \cdot \delta_2 \cdot \alpha^2 + \cdots + 10\, \delta_1 \cdot \alpha + \cdots
\end{aligned}
}{
\begin{aligned}
& 10000\, \beta^2 \cdot \alpha^2 - 20000\, \beta \cdot p_5 \cdot \alpha^2 + \cdots + \\
& 200\, \beta^2 \cdot p_5^2 \cdot p_2 + \cdots + 40\, p_2 + 1
\end{aligned}
}
\label{eq:symbolic_s7_simplified}
\end{equation}

\noindent The full symbolic expression, including all
terms and recursive retry contributions up to $p_7^{40}$, is included in Appendix~B. These terms capture nested recovery paths involving both human and autonomous strategies, enabling quantitative comparison of mitigation effectiveness across varying parameter settings.

\begin{figure}
    \centering
    \includegraphics[width=0.48\textwidth]{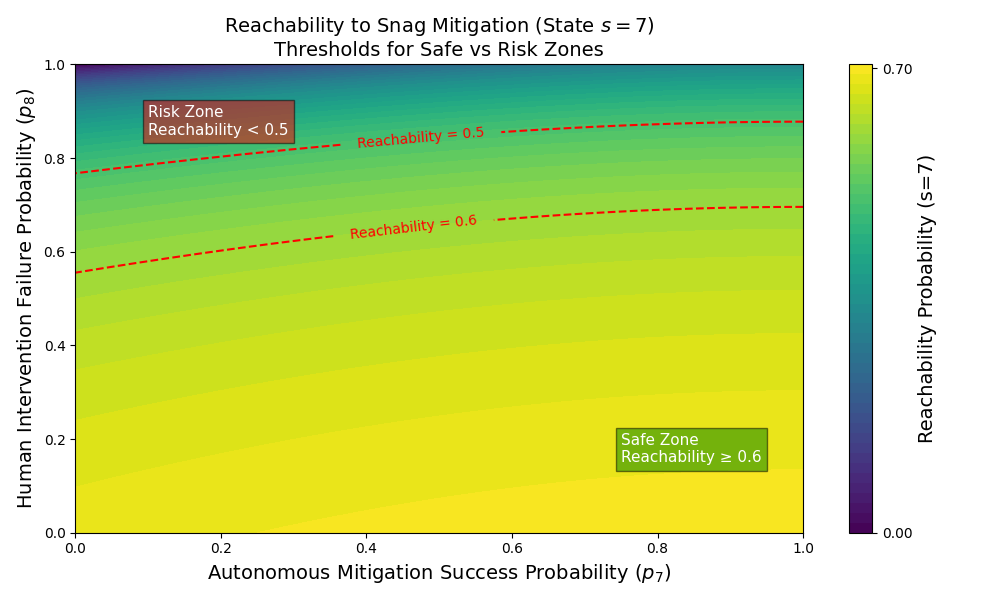}
    \caption{Probability of successful mitigation (reaching $s_7$) for varying autonomous mitigation success probability ($p_7$) and human failure probability ($p_8$). Lighter regions indicate higher success rates.}
    \label{fig:mitigationSnag}

\end{figure}

Figure~\ref{fig:mitigationSnag} shows the evaluated symbolic expression over a range of human and autonomous recovery capabilities. A safety threshold contour at $P=0.5$ distinguishes high-risk scenarios (top-left) from safer operating regions (bottom-right). The \textit{Safe Zone} is defined where the probability of reaching sate $s=7$ (snag mitigation) exceeds 0.6. This typically occurs when the autonomous mitigation success probability ($p_7$) is high, or the human intervention failure probability ($p_8$) is low. 

The results highlight an important compensation effect: strong performance in one recovery pathway (human or autonomous) can offset weaknesses in the other. For example, even when $p_7$ is low, maintaining a low $p_8$ (i.e., reliable human intervention) preserves high overall success rates. Conversely, increasing $p_7$ improves robustness against human failure. 

These finding reinforce the benefit of hybrid, adaptive recovery strategies in which human-in-the-loop support and autonomous recovery dynamically compensate for each other. The symbolic reachability analysis thus provides both an interpretable quantitative basis for design decisions and an analytical foundation for runtime adaptation.

\subsection{Symbolic Analysis of Snag Escalation Failure}\label{sub:eval-symbEscalation}

We now analyse the second key reachability property of interest $P_{=?} [ F\ s=2]$. This property quantifies the probability of reaching state $s_2$, representing an undetected escalation of a detected snag — a critical failure scenario in the robot-assisted dressing task.  

Using \texttt{PRISM+PARAM}, we symbolically evaluated this property while varying two key parameters: (1) the probability of detecting a potential snag, $p_2$,
and (2) the probability of remaining in the monitoring state without triggering mitigation, $p_3$. These
directly govern the likelihood of silent escalation, with all other transitions held constant to isolate their effect. 

The symbolic expression extracted from \texttt{PRISM+PARAM} was algebraically simplified and normalised. It captures both direct and indirect contributions to $s_2$ reachability via recursive monitoring cycles. A normalised version
was evaluated over a grid of $p_2$ and $p_3$ values, revealing the influence of detection and monitoring performance.

The simplified symbolic expression is given by:

\begin{equation}
P = \frac{100\,P_3 \cdot P_2 + 98\,P_2 - 99}{88\,P_2 - 100}
\label{eq:reachability_s2_simplified}
\end{equation}

\begin{figure}
    \centering
    \includegraphics[width=0.48\textwidth]{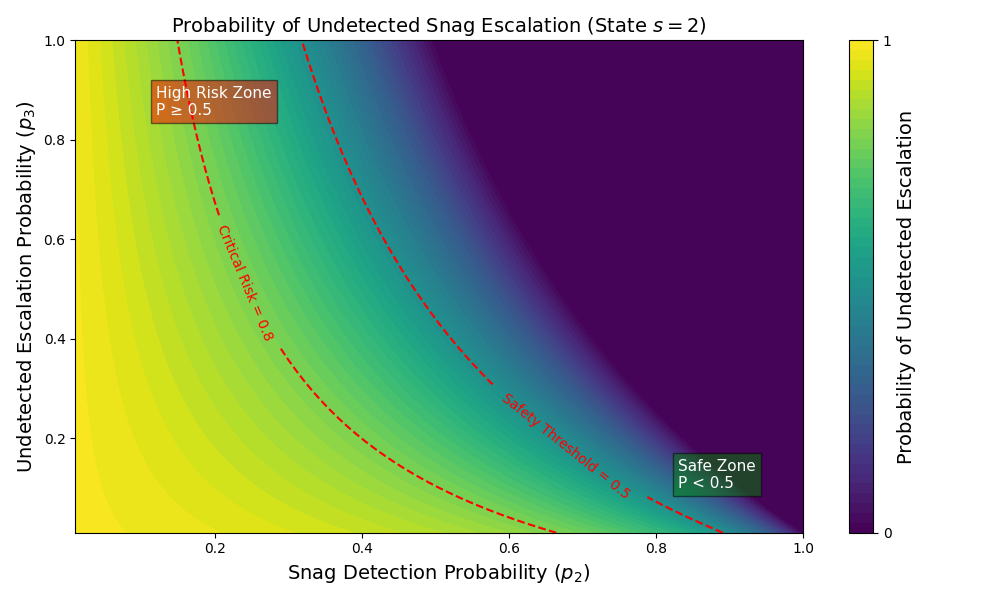}
    \caption{Reachability probability of undetected snag escalation ($s_2$) as a function of detection probability $p_2$ and undetected escalation probability $p_3$. Lighter regions indicate higher failure probability. Dashed contour lines highlight key risk thresholds: $P=0.5$ (Safety Threshold) and $P=0.8$ (Critical Risk). Annotated zones mark high-risk and safe regions.}
    \label{fig:snag_escalation_heatmap}

\end{figure}

\noindent This
captures the 
trade-off between detection ($p_2$) and monitoring ($p_3$). The
probability increases linearly with both $p_2$ and $p_3$ in the numerator, while the denominator introduces a non-linear effect governed solely by $p_2$. As $p_2$ increases (detection improves), the denominator approaches zero from below, causing a steep drop in failure probability — a behaviour clearly reflected in the heatmap gradient (Figure~\ref{fig:snag_escalation_heatmap}). 

Figure~\ref{fig:snag_escalation_heatmap} illustrates the reachability landscape for undetected snag escalation, based on the symbolic property $P_{=?}[F \text{ s=2 }]$. The heatmap shows how the failure probability varies over a grid of snag detection probabilities ($p_2$) and escalation persistence probabilities ($p_3$). 

The \textit{``Safe Zone"} ($P<0.5$) occurs in the lower-right corner, where detection is strong ($p_2 \rightarrow 1$) and escalation is unlikely ($p_3 \rightarrow 0$). The \textit{``High Risk Zone"} ($P\ge 0.5$) appears in the upper-left corner, where detection is weak ($p_2 \rightarrow 0$) and escalation is persistent ($p_3 \rightarrow 1$). 

The red dashed contours highlight two thresholds: the safety threshold at $P=0.5$, and the critical risk boundary at $P=0.8$, making conditions where the likelihood of silent failure is dangerously high. Notably, the steep gradient along the $p_2$ axis reveals that even modest gains in detection capability can substantially reduce failure risk. 

These findings highlight the system's high sensitivity to snag detection accuracy and provide actionable insights for both formal verification and runtime adaptation. Prioritising high $p_2$ values can significantly lower the probability of silent escalation, thereby improving overall safety.

\subsection{Proxy Cost Analysis of Undetected Snag Escalation}\label{sec:cost_analysis}

To evaluate the cost implications of undetected snag escalation, we adopt a proxy formulation based on the symbolic reachability expression for state $s=2$ (undetected escalation). Specifically, we compute Equation~\ref{eq:ExpectedCost_Fs=2}, where $C_{S2}$ is a fixed penalty associated with entering the failure state. This formulation provides a lightweight approximation of the cost impact by scaling the symbolic reachability with the defined failure cost. It enables efficient evaluation during runtime and captures the increasing cost of failure under unsafe parameter regimes.

\begin{figure}
    \centering
    \includegraphics[width=0.48\textwidth]{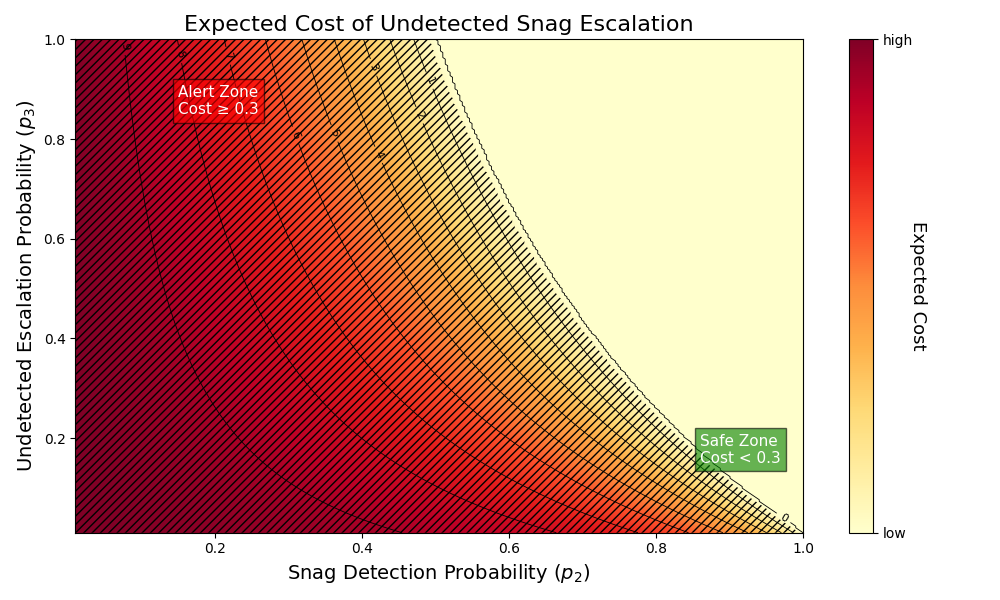}
    \caption{Annotated heatmap showing the expected cost of undetected snag escalation, computed as $C_{S2} \cdot P_{=?}[F\, s=2]$, over varying detection probability ($p_2$) and escalation persistence ($p_3$). The \textit{Alert Zone} (Cost $\geq$ 0.3) highlights failure-prone conditions; the \textit{Safe Zone} (Cost $<$ 0.3) reflects robust detection and recovery.}
    \label{fig:expected_cost_s2}

\end{figure}

Figure~\ref{fig:expected_cost_s2} visualises this cost landscape. The \textit{``Alert Zone"} ($\text{Cost}\ge0.3$) arises when snag detection is weak ($p_2 \rightarrow 0$) and the likelihood of persistent, undetected escalation is high ($p_3 \rightarrow 1$). The hatched area clearly marks this high-risk region. 

Conversely, the \textit{``Safe Zone"} ($\text{Cost}<0.3$) occupies the lower-right corner of the heatmap, where effective detection significantly reduces escalation risk and associated penalty. 

Compared to the raw reachability view in Figure~\ref{fig:snag_escalation_heatmap}, this proxy formulation adds interpretability by contextualising risk in cost terms. This allows the system to perform low-overhead runtime checks against unsafe conditions using a single symbolic expression and supports proactive adaptation in safety-critical human-robot interactions.

\subsection{Symbolic Reward Analysis of Mitigation Success}\label{sec:s7-reward}

This analysis focuses on the expected reward associated with successful snag mitigation, represented by reaching state $s_7$ in the pDTMC model. Since cumulative reward properties cannot currently be symbolically evaluated in \texttt{PRISM+PARAM}, we compute the expected reward using a scaled reachability formulation from Equation~\ref{eq:ExpectedReward_Fs=7}, where \(R_{S7}\) is a fixed reward assigned to successful mitigation. The symbolic reachability expression \(P_{=?}[F\, s=7]\) was derived in Section~\ref{sub:eval-symbReachability} and captures the combined effect of detection, escalation, and mitigation strategies (both human and autonomous).

To examine reward trade-offs, we fix parameters \(p_2 = 0.9\), \(p_4 = 0.7\), and \(p_5 = 0.65\), and vary:

\begin{itemize}
    \item \(p_8\): the failure probability of human intervention,
    \item \(p_7\): the success probability of autonomous recovery.
\end{itemize}

The symbolic expression is evaluated over this 2D parameter space, and the expected reward is computed accordingly.

\begin{figure}
    \centering
    \includegraphics[width=0.48\textwidth]{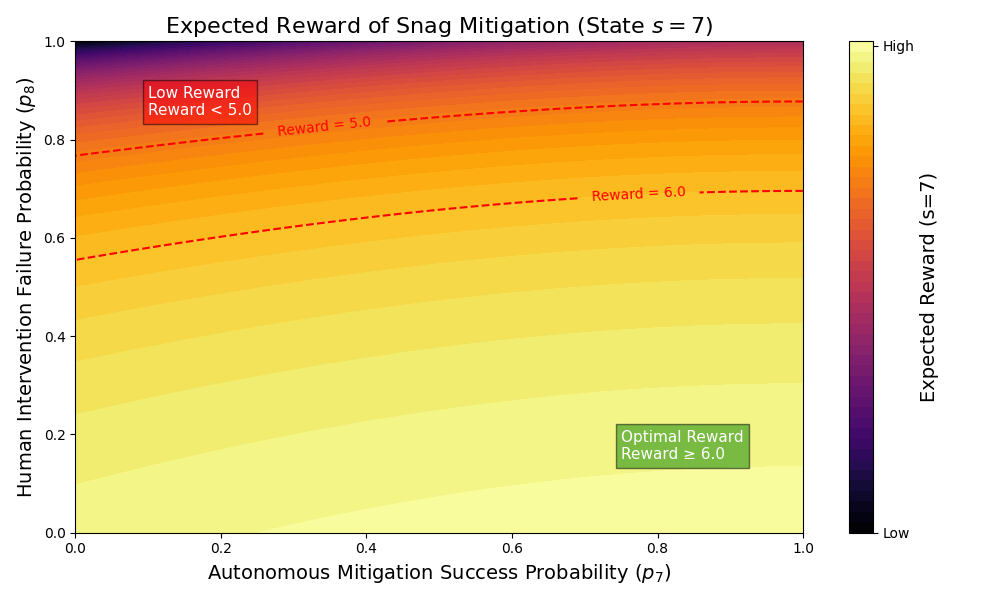}
    \caption{Expected reward for mitigation success (\(s=7\)) evaluated from the symbolic expression over human intervention failure (\(p_8\)) and autonomous recovery success (\(p_7\)). The \textit{Optimal Reward} zone arises when both modalities are reliable; \textit{Low Reward} emerges when both pathways are likely to fail.}
    \label{fig:s7-reward}
\end{figure}

Figure~\ref{fig:s7-reward} illustrates the resulting expected reward surface. A clear \textit{``Optimal Reward"} zone emerges in the bottom-right corner ($p_8 \rightarrow 0$, $p_7 \rightarrow 1$), where both mitigation pathways are highly effective. In contrast, the \textit{``Low Reward"} region appears when both autonomous and human strategies are likely to fail ($p_7 \rightarrow 0$, $p_8 \rightarrow 1$). Contour lines at reward thresholds $R = 5.0$ (Lower Bound) and $R = 6.0$ (Optimal Reward) delineate these regions.

These findings reinforce the hybrid design principle: strong performance in either the human-in-the-loop or autonomous recovery pathway can compensate for the other’s limitations, enabling robust, reward-aware adaptation in safety-critical scenarios.


\section{Discussion}
\label{sec:discussion}

Our approach demonstrates how symbolic verification, Bayesian adaptation, and hazard-informed modelling can be combined to support safe, adaptive decision-making in human-robot collaboration. While developed for robot-assisted dressing (RAD), the methodology is applicable to broader pHRI domains where runtime uncertainty and task variability are critical. 
The symbolic evaluation presented in Section~\ref{sec:evaluation} shows that requirements like snag escalation and mitigation success can be interpreted as functions of runtime-updated parameters such as $p_2$, $p_3$, $p_7$, and $p_8$. These relationships allow the controller to identify and avoid unsafe regions of the parameter space proactively.

A key benefit is that runtime verification does not require re-running a model checker. Once symbolic expressions are precomputed, parameter substitution enables real-time evaluation with minimal computational overhead. This facilitates explainable decision-making and supports runtime assurance. 
One current limitation is that symbolic evaluation is based on an abstract model that does not fully capture the continuous dynamics of low-level controllers. Further work will integrate these physical dynamics to bridge the gap between symbolic guarantees and physical execution. 


\section{Conclusion and Future Work}
\label{sec:conclusion}
This paper introduced a high-level control framework for robot-assisted dressing that integrates symbolic verification with runtime adaptation. By modelling the dressing task using a pDTMC and deriving closed-form expressions for key PCTL requirements, our framework enables fast runtime evaluation without re-invoking a model checker. Bayesian inference is used to update transition probabilities from real-time sensory and user feedback. These estimates are substituted into precomputed expressions to assess reachability, cost, and reward trade-offs. The resulting analysis guides safe and adaptive decision-making based on evolving risk profiles. 

Future work includes tighter integration between the symbolic model and low-level robot control dynamics, enabling end-to-end assurance across the stack. 
Finally, given the complexity of the robot-assisted dressing task, in this paper, we focus exclusively on the snagging mitigation strategy. To advance towards an fully assured RAD system will require modelling a wider range of scenarios that impact the overall safety. To that end, we will investigate the development of a RAD multi-model world, following the approach in~\cite{calinescu2025verification}, to support the verification of system-level requirements.

\section*{Acknowledgment}

This work was funded by the EPSRC projects EP/V026747/1 `UKRI Trustworthy Autonomous Systems Node in Resilience’ and EP/V026801/2 `UKRI Trustworthy Autonomous Systems Node in Verifiability’.

\bibliographystyle{IEEEtran}
\bibliography{ref}

\onecolumn 
\section*{Appendix A: PRISM DTMC Model for Snagging Scenarios}
\label{appendix:dtmc}

The following code represents the updated PRISM DTMC model for snagging scenarios in the Robot-Assisted Dressing (RAD) system. This version matches the symbolic model used in the reachability, cost, and reward analyses discussed in the main paper.

\begin{lstlisting}[language=PRISM, caption={Updated PRISM DTMC Model for Snagging Scenarios}, label={lst:prism_model}]
dtmc

// Constants
const double C_S2;
const double C_S8;
const double R_S7;
const double BASE_REWARD_S3;
const double P1 = 0.4;
const double P2;
const double P3;
const double P4;
const double P5;
const double P6;
const double P7;
const double P8;
const double P9;
const double p10;
const int MAX_TIME_TRAJECTORY;
const int MAX_TIME;

// State module
module robot_assisted_dressing
    s : [0..9] init 0;
    t : [0..MAX_TIME] init 0;
    time_step : [0..MAX_TIME] init 0;
    trajectory_complete : bool init false;

    // S0 transitions
    [] s=0 & time_step < MAX_TIME_TRAJECTORY & !trajectory_complete ->
        P9: (s'=0) & (time_step'=time_step+1) +
        P2: (s'=1) & (time_step'=time_step+1) +
        (1 - (P2 + P9)): (s'=2) & (time_step'=time_step+1);
    [] s=0 & time_step = MAX_TIME_TRAJECTORY ->
        1.0: (s'=3) & (trajectory_complete'=true);

    // S1 transitions
    [] s=1 & time_step < MAX_TIME_TRAJECTORY & !trajectory_complete ->
        P3: (s'=1) & (time_step'=time_step+1) +
        P4: (s'=4) & (time_step'=time_step+1) +
        (1 - (P3 + P4)): (s'=2) & (time_step'=time_step+1);
    [] s=1 & time_step = MAX_TIME_TRAJECTORY ->
        1.0: (s'=3) & (trajectory_complete'=true);

    // S2: undetectedEscalation
    [] s=2 -> 1.0: (s'=8) & (time_step'=0);

    // S3: dressingComplete
    [] s=3 -> 1.0: (s'=9);

    // S4: mitigation decision
    [] s=4 -> 
        P5: (s'=5) +
        P6: (s'=6) +
        (1 - (P5 + P6)): (s'=8);

    // S5: HRI
    [] s=5 -> 
        P8: (s'=4) +
        (1 - P8): (s'=7);

    // S6: Autonomous mitigation
    [] s=6 & t < MAX_TIME ->
        P7: (s'=7) +
        (1 - P7): (s'=6) & (t'=t+1);
    [] s=6 & t = MAX_TIME ->
        1.0: (s'=8);

    // S7: mitigationSuccess
    [] s=7 -> 1.0: (s'=0) & (time_step'=0);

    // S8: abortTask
    [] s=8 -> 1.0: (s'=9) & (time_step'=0);

    // S9: terminal
    [] s=9 -> 
        p10: (s'=9) +
        (1 - p10): (s'=0);
endmodule

// Rewards
rewards "time"
    true : 1;
endrewards

rewards "cost_s2"
    s=2 : C_S2;
endrewards

rewards "cost_s8"
    s=8 : C_S8;
endrewards

rewards "reward_s7"
    s=7 : R_S7;
endrewards

rewards "reward_s3"
    s=3 : BASE_REWARD_S3;
endrewards
\end{lstlisting}

\vspace{1em}
\noindent
\textbf{Tool Configuration for Symbolic Evaluation.} To symbolically evaluate the two reachability requirements discussed in Sections~\ref{sub:eval-symbReachability} and~\ref{sub:eval-symbEscalation}, we used the \texttt{PRISM+PARAM} command-line interface with the model in Listing~\ref{lst:prism_model}. 
Parametrised expressions were generated using the following configurations:

\begin{itemize} \item \textbf{Mitigation Success ($P=?[F, s=7]$):} \begin{verbatim} prism pDTMC_snag_cost_integrated_time8.pm snagDtmc_p1.pctl
-param 'P2=0.06:0.07,P4=0.087:0.88,P5=0.6:0.7,P7=0.07:0.8,P8=0.04:0.05'
-const 'P3=0.1,P6=0.05,P9=0.1,p10=0.8,C_S2=10,C_S8=5,R_S7=10,
BASE_REWARD_S3=20,MAX_COST=100,MAX_TIME_TRAJECTORY=2,MAX_TIME=2' \end{verbatim}

\item \textbf{Undetected Snag Escalation ($P=?[F\, s=2]$):}
\begin{verbatim}
prism pDTMC_snag_cost_integrated_time8.pm snagDtmc_p2.pctl \
      -param 'P2=0.06:0.07,P3=0.05:0.1' \
      -const 'P4=0.88,P5=0.7,P6=0.05,P7=0.8,P8=0.05,P9=0.1,p10=0.8,\
              C_S2=10,C_S8=5,R_S7=10,BASE_REWARD_S3=20,\
              MAX_COST=100,MAX_TIME_TRAJECTORY=2,MAX_TIME=2'
\end{verbatim}
\end{itemize}
\noindent
These configurations enabled symbolic exploration of key parameter ranges while keeping less influential constants fixed for tractability. The resulting expressions (see Appendix~B) were then used in our analysis and visualisation scripts.

\section*{Appendix B: Symbolic Reachability Expression for \( P = ? [F\, \text{s=7}] \)}
\label{appendix:symbExpSuccesfulMitigation}

The symbolic expression for the reachability probability \( P = ? [F\, \text{s=7}] \)—representing the probability of successful snag mitigation—was generated using the \texttt{PRISM+PARAM} tool. This tool performs symbolic parametric model checking over a discrete-time Markov chain (pDTMC), producing closed-form expressions that capture the impact of transition probabilities on key system requirements.

In this case, the expression incorporates five parameters: \( p_2 \), \( p_4 \), \( p_5 \), \( p_7 \), and \( p_8 \). These correspond to snag detection, escalation, initiation of mitigation, autonomous success, and human intervention failure, respectively.

\subsection*{Full Rational Expression}

The symbolic reachability probability of reaching state \( s=7 \) is given by the following rational expression:

\begin{equation}
P_{mitigated} = \frac{\text{Numerator Polynomial}}{\text{Denominator Polynomial}}
\end{equation}

\noindent where:

\textbf{Numerator Polynomial}:
\begin{multline*}
10000\, p_8^2 p_5^2 p_4^2 p_2^2 - 20000\, p_8 p_5^2 p_4^2 p_2^2 + 2000\, p_8^2 p_5^2 p_4 p_2^2 + 100\, p_8 p_7^2 p_5 p_4 p_2^2 \\
+ 10000\, p_5^2 p_4^2 p_2^2 - 500\, p_8 p_5 p_4^2 p_2^2 - 2000\, p_8 p_5^2 p_4 p_2^2 - 200\, p_8 p_7 p_5 p_4 p_2^2 \\
+ 100\, p_8^2 p_5^2 p_4 p_2 + 5\, p_8 p_7^2 p_5 p_4 p_2 + 500\, p_5 p_4^2 p_2^2 - 2000\, p_8 p_5 p_4 p_2^2 \\
- 100\, p_7^2 p_4 p_2^2 - 100\, p_8 p_5^2 p_4 p_2 - 10\, p_8 p_7 p_5 p_4 p_2 + 2000\, p_5 p_4 p_2^2 \\
+ 200\, p_7 p_4 p_2^2 - 100\, p_8 p_5 p_4 p_2 - 5\, p_7^2 p_4 p_2 + 100\, p_5 p_4 p_2 + 10\, p_7 p_4 p_2
\end{multline*}

\textbf{Denominator Polynomial}:
\begin{multline*}
10000\, p_8^2 p_5^2 p_4^2 p_2^2 - 20000\, p_8 p_5^2 p_4^2 p_2^2 + 4000\, p_8^2 p_5^2 p_4 p_2^2 + 10000\, p_5^2 p_4^2 p_2^2 \\
- 500\, p_8 p_5 p_4^2 p_2^2 - 4000\, p_8 p_5^2 p_4 p_2^2 + 400\, p_8^2 p_5^2 p_2^2 + 200\, p_8^2 p_5^2 p_4 p_2 \\
+ 500\, p_5 p_4^2 p_2^2 - 4100\, p_8 p_5 p_4 p_2^2 - 200\, p_8 p_5^2 p_4 p_2 + 40\, p_8^2 p_5^2 p_2 \\
+ 4000\, p_5 p_4 p_2^2 - 800\, p_8 p_5 p_2^2 - 205\, p_8 p_5 p_4 p_2 + p_8^2 p_5^2 + 100\, p_4 p_2^2 \\
+ 200\, p_5 p_4 p_2 - 80\, p_8 p_5 p_2 + 400\, p_2^2 + 5\, p_4 p_2 - 2\, p_8 p_5 + 40\, p_2 + 1
\end{multline*}

\subsection*{Symbolic Interpretation and Substitution}

To support runtime substitution and improve interpretability, we define the following symbolic terms capturing meaningful joint behaviours:

\begin{itemize}
    \item \( \alpha = p_2 \cdot p_4 \): Joint probability of snag detection and escalation.
    \item \( \beta = p_5 \cdot (1 - p_8) \): Effective probability of successful human intervention.
    \item \( \delta = p_5 \cdot p_8 \cdot p_7 \): Autonomous recovery success following human failure.
    \item \( \epsilon = p_7 \), \( \zeta = p_7^2 \): Contributions from autonomous retry attempts.
\end{itemize}

These symbolic terms appear in nested and cross-multiplied combinations in the full expression, reflecting the interaction between parallel recovery pathways and retry dynamics.

\subsection*{Interpretation}

This symbolic reachability expression for \( P = ? [F\, s=7] \) captures several key dynamics:

\begin{itemize}
    \item The effect of early detection and escalation through \( \alpha \),
    \item The role of both successful and failed human interventions via \( \beta \) and \( \delta \),
    \item The increasing influence of autonomous recovery through terms in \( p_7 \) and \( p_7^2 \),
    \item The interaction of parameters in both numerator and denominator, modelling compensatory behaviours (e.g., strong autonomy offsets poor human response).
\end{itemize}

Despite its complexity, the structure of this expression enables direct substitution of runtime estimates for each parameter, making it practical for adaptive control and online verification without re-running model checking.

\end{document}